\pdfoutput=1

\documentclass[11pt]{article}

\usepackage{acl}

\usepackage{times}
\usepackage{latexsym}
\usepackage{graphicx}

\usepackage[T1]{fontenc}

\usepackage[utf8]{inputenc}

\usepackage{microtype}
\usepackage{multicol}
\usepackage{amsmath}
\usepackage{amsfonts,amssymb} 
\usepackage{booktabs}
\usepackage[normalem]{ulem}
\useunder{\uline}{\ul}{}
\usepackage{multirow}
\usepackage{booktabs}
\usepackage{float}
\usepackage{pifont}
\usepackage{array}
\usepackage{color}
\usepackage[ruled]{algorithm2e}
\usepackage{pgfplots} 
\pgfplotsset{compat=1.18} 

\def\mystrut{\vphantom{hg}}

\pgfplotsset{
    legend image with text/.style={
        legend image code/.code={%
            \node[anchor=center] at (0.3cm,0cm) {#1};
        }
    },
    xticklabel style={/pgf/number format/fixed}, 
}

\title{Adversarial Multi-task Learning for End-to-end Metaphor Detection}

\author{Shenglong Zhang \quad {Ying Liu \Thanks{ Corresponding Author}} \\ Tsinghua University, Beijing, China, 100084 \\ \texttt{zsl18@mails.tsinghua.edu.cn} \\ \texttt{yingliu@mail.tsinghua.edu.cn}}

\begin{document}
\maketitle

\begin{abstract}
Metaphor detection (MD) suffers from limited training data. In this paper, we started with a linguistic rule called Metaphor Identification Procedure and then proposed a novel multi-task learning framework to transfer knowledge in basic sense discrimination (BSD) to MD. BSD is constructed from word sense disambiguation (WSD), which has copious amounts of data. We leverage adversarial training to align the data distributions of MD and BSD in the same feature space, so task-invariant representations can be learned. To capture fine-grained alignment patterns, we utilize the multi-mode structures of MD and BSD. Our method is totally end-to-end and can mitigate the data scarcity problem in MD. Competitive results are reported on four public datasets. Our code and datasets are available \footnote{https://github.com/SilasTHU/AdMul}.
\end{abstract}

\section{Introduction}

Metaphor involves a mapping mechanism from the source domain to the target domain, as proposed in Conceptual Metaphor Theory \citep{lakoff2008metaphors}. 

e.g. \ \emph{The police \textbf{smashed} the drug ring after they were tipped off .}

\textbf{\textit{Smash}} in the above sentence means "hit hard" literally (source domain). However, it is employed in a creative way, indicating "overthrow or destroy" (target domain). The mapping from the source to the target makes the word a metaphor.

Understanding metaphors in human languages is essential for a machine  to dig out the underlying intents of speakers. Thus, metaphor detection and understanding are crucial for sentiment analysis \cite{8267597}, and machine translation\citep{ mao-etal-2018-word}, etc.

Metaphor detection (MD) requires a model to predict whether a specific word is literal or metaphorical in its current context. Linguistically, if there is a semantic conflict between the contextual meaning and a more basic meaning, the word is a metaphor \citep{crisp2007mip, steen2010method, do-dinh-etal-2018-weeding}. The advent of large Pre-trained Language Models has pushed the boundaries of MD far ahead \citep{devlin-etal-2019-bert, https://doi.org/10.48550/arxiv.1907.11692}. However, MD suffers from limited training data, due to complex and difficult expert knowledge for data annotation \citep{group2007mip}.

Recently, \citet{lin-etal-2021-cate} used self-training to expand MD corpus, but error accumulation could be a problem. Many researchers used various external knowledge like part of speech tags \citep{su-etal-2020-deepmet, choi-etal-2021-melbert}, dictionary resources \citep{su-etal-2021-enhanced, zhang-liu-2022-metaphor}, dependency parsing \citep{Le_Thai_Nguyen_2020, song-etal-2021-verb}, etc., to promote MD performance. These methods are not end-to-end, thus they impeded continuous training on new data.

To address the data scarcity problem in MD, we propose a novel task called basic sense discrimination (BSD) from word sense disambiguation (WSD). BSD regards the most commonly used lexical sense as a basic usage, and aims to identify whether a word is basic in a certain context. Both BSD and MD need to compare the semantic difference between the basic meaning and the current contextual meaning. Despite the lack of MD data, we can distill knowledge from BSD to alleviate data scarcity and overfitting, which leads to the usage of multi-task learning.


We design the \textbf{Ad}versarial \textbf{Mul}ti-task Learning Framework (AdMul) to facilitate the knowledge transfer from BSD to MD. AdMul aligns the data distributions for MD and  BSD through adversarial training to force shared encoding layers (for example, BERT) to learn task-invariant representations. Furthermore, we leverage the internal multi-mode structures for fine-grained alignment. The literal senses in MD are forcibly aligned with basic senses in BSD, which can push the literal senses away from the metaphorical. Similarly, the non-basic senses in BSD are aligned with metaphors in MD, which enlarges the discrepancy between basic and non-basic senses to enhance model performance.

The contributions of this paper can be summarized as follows:

\begin{itemize}
\item[$\bullet$] We proposed a new task, basic sense discrimination, to promote the performance of metaphor detection via a multi-task learning method. The data scarcity problem in MD can be mitigated via knowledge transfer from a related task.
\item[$\bullet$] Our proposed model, AdMul, uses adversarial training to learn task-invariant representations for metaphor detection and basic sense discrimination. We also make use of multi-mode structures for fine-grained alignment. Our model is free of any external resources, totally end-to-end, and can be easily trained.
\item[$\bullet$] Experimental results indicate that our model achieves competitive performance on four datasets due to knowledge transfer and the regularization effect of multi-task learning. Our zero-shot transfer result even surpasses fine-tuned baseline models.
\end{itemize}

\section{Related Work}
\noindent \textbf{Metaphor Detection: } Metaphor detection is a popular task in figurative language computing \citep{leong-etal-2018-report, leong-etal-2020-report}. With the progress of natural language processing, various methods have been proposed. Traditional approaches used different linguistic features like word abstractness, word concreteness, part of speech tags and linguistic norms, etc., to detect metaphors \citep{shutova-sun-2013-unsupervised, tsvetkov-etal-2014-metaphor, beigman-klebanov-etal-2018-corpus, wan-etal-2020-using}. These methods are not end-to-end and rely heavily on feature engineering.

The rise of deep learning boosted the advancement of metaphor detection significantly. \citet{gao-etal-2018-neural}, \citet{wu-etal-2018-neural} and \citet{mao-etal-2019-end} used RNN and word embeddings to train MD models. Recently, lots of works combined the advantages of pre-trained language models and external resources to enhance the performance of metaphor detection \citep{su-etal-2020-deepmet, su-etal-2021-enhanced, choi-etal-2021-melbert, song-etal-2021-verb, zhang-liu-2022-metaphor}. Though great improvements have been made, these models still suffer from the lack of training data, which is well exemplified by their poorer performance on small datasets.

\noindent \textbf{Multi-task Learning: } Multi-task learning (MTL) can benefit a target task via related tasks. It has brought great success in computer vision and natural language processing. MTL learns universal representations for different task inputs, so all tasks share a common feature space, where knowledge transfer becomes possible. Previous studies trained MTL models by deep neural networks like CNN or RNN, achieving promising results in text classification \citep{liu-etal-2017-adversarial, chen-cardie-2018-multinomial}. \citet{liu-etal-2019-multi} and \citet{clark-etal-2019-bam} combined MTL framework with BERT \citep{devlin-etal-2019-bert}, obtaining encouraging results on multiple GLUE tasks. There are some other successful MTL applications in machine translation \citep{dong-etal-2015-multi}, information extraction \citep{nishida-etal-2019-answering}, and sentiment analysis \citep{liang-2020}, etc. \citet{dankers-etal-2019-modelling} applied MTL to study the interplay of metaphor and emotion. \citet{Le_Thai_Nguyen_2020} combined WSD and MD for better metaphor detection results. However, to the best of our knowledge, we are the first to use adversarial MTL for metaphor detection based on the linguistic nature of metaphors.

\begin{figure*}[ht]
\centering
\includegraphics[width=\textwidth]{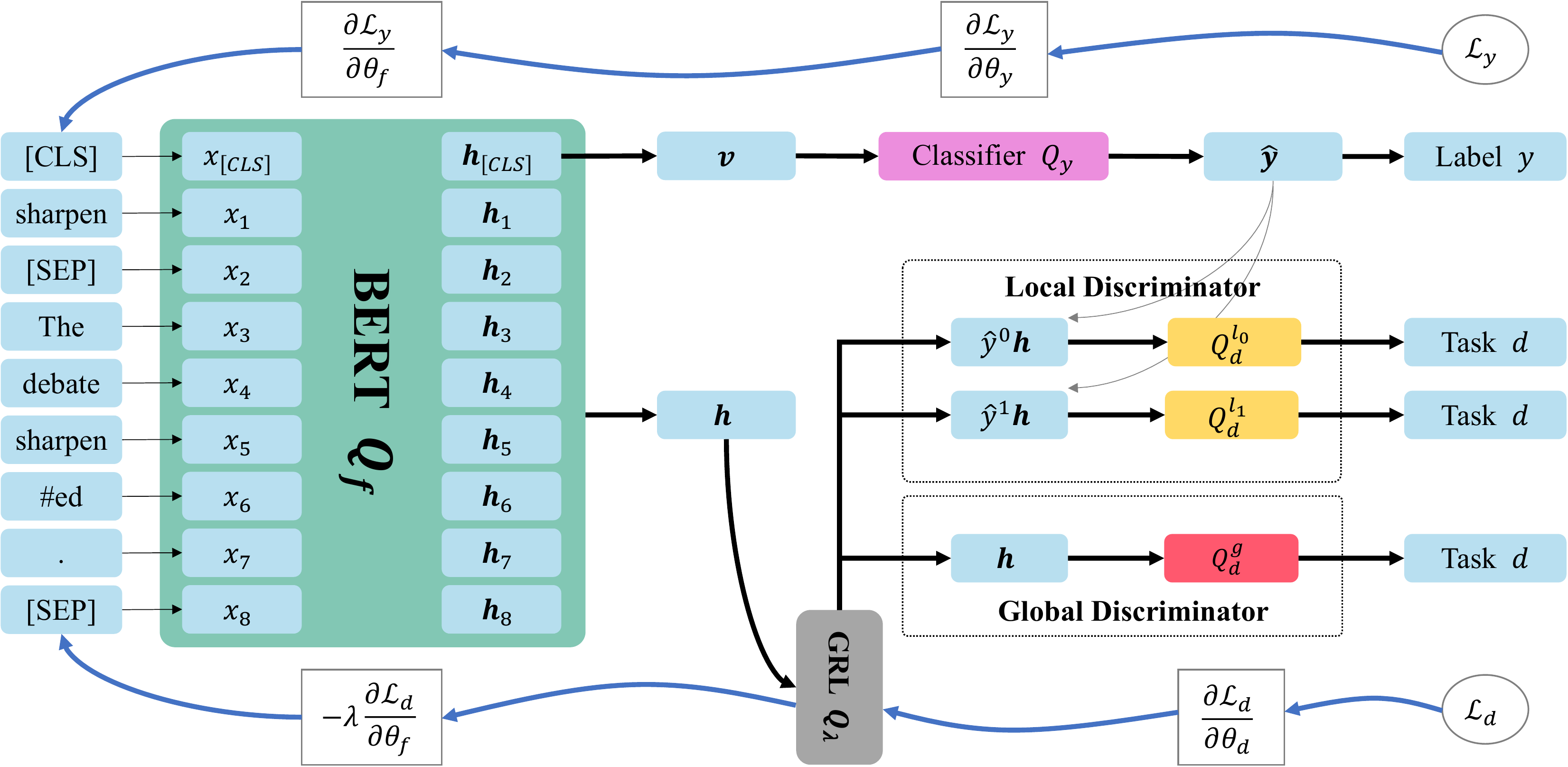}
\caption{AdMul architecture. The \textbf{black} arrows mean forward propagation, while the \textcolor[RGB]{68,114,196}{\textbf{blue}} ones denote back propagation. BERT is the shared feature extractor $Q_f$. GRL stands for gradient reversal layer. Classifier $Q_y$ is task-specific to perform MD or BSD, and $y$ is the label for MD or BSD. Global discriminator $Q_d^g$ aligns overall data distribution to make BERT learn universal representations. Two local discriminators $Q_d^{l_c}$ are in line with two labels. Each is responsible for aligning the data in MD and BSD of label $c$. Both $Q_d^g$ and $Q_d^{l_c}$ predict which task the input sentence comes from. Task $d \in \{0, 1\}$, 0 or MD and 1 for BSD. $\mathcal{L}_y$ is the loss for $Q_y$. $\mathcal{L}_d$ is the loss for $Q_d^g$ or $Q_d^{l_c}$.}
\label{model}
\end{figure*}

\section{Proposed Method}

\subsection{Metaphor Identification Procedure}
Metaphor Identification Procedure (MIP) \citep{crisp2007mip} is the most commonly used linguistic rule in guiding metaphor detection. It is originally the construction guideline of VU Amsterdam Metaphor Corpus. MIP indicates that if a word contrasts with one of its more basic meanings but can be understood in comparison with it, then the word is a metaphor. A more basic meaning is more concrete, body-related, more precise, or historically older \citep{steen2010method, do-dinh-etal-2018-weeding}.

Some researchers have pointed out that when a word is used alone, it is very likely to depict a more basic meaning \citep{choi-etal-2021-melbert, song-etal-2021-verb}. We concatenate the target word and the sentence as input. In the input, the first segment is the target used alone, presenting a more basic meaning. The second segment is the whole sentence, which can encode the contextual meaning of the target. Then the model adopts MIP to detect metaphors.

\subsection{From WSD to BSD}
Metaphor detection (MD) aims to identify whether a contextualized word is metaphorical. Word sense disambiguation (WSD) aims to determine the lexical sense of a certain word from a given sense inventory. The two tasks share the same nature: we should decide the sense of a given word according to its context.

A word may have multiple senses, so WSD is a multinomial classification task, whereas MD is a binary classification task. Integrating WSD with MD can be quite expensive. For example, the state-of-the-art model \citep{barba-etal-2021-consec} regarded WSD as an information extraction task. It concatenated all the candidate senses and tried to extract the correct one. Such a method requires not only external dictionary resources, but also enormous computing resources since the input may be a very long sequence.

WordNet \citep{wordnet1, wordnet2} ranks the senses of a word according to its occurrence frequency\footnote{https://wordnet.princeton.edu/frequently-asked-questions}. The most commonly used lexical sense is at the top of the inventory list, which is usually a more basic meaning\citep{choi-etal-2021-melbert, song-etal-2021-verb, zhang-liu-2022-metaphor}. Thus, we regard the most commonly used sense as a basic sense of a word, and try to figure out whether a word in a certain context is basic or not. We call this task basic sense discrimination (BSD). Obviously, BSD is a binary classification task and fits MD.

\subsection{Task Description}
Formally, given the MD dataset $\mathcal{D}_{\text{MD}} = \{{(\boldsymbol{x}_i^{\text{MD}}\, y_i^{\text{MD}})}_{i=1}^{n_\text{MD}}\}$ and the BSD dataset $\mathcal{D}_{\text{BSD}} = \{{(\boldsymbol{x}_i^{\text{BSD}}, y_i^{\text{BSD}})_{i=1}^{n_\text{BSD}}}\}$, they have $n_{\text{MD}}$ and $n_{\text{BSD}}$ labeled training samples respectively. $\boldsymbol{x} = \left( \text{[CLS]}, \text{target}, \text{[SEP]}, \text{sentence}, \text{[SEP]}\right )$. Usually, MD and BSD have different data distributions $p$, so $p_{\text{MD}}(\boldsymbol{x}_{\text{MD}}) \neq p_{\text{BSD}}(\boldsymbol{x}_{\text{BSD}})$. Both $\mathcal{D}_{\text{MD}}$ and$\mathcal{D}_{\text{BSD}}$ will be used to train a multi-task learning model, which will align $p_{\text{MD}}$ and $p_{\text{BSD}}$ in a same feature space via adversarial training. Our goal is to minimize the risk $\epsilon = \mathbb{E}_{(\boldsymbol{x}, y)\sim p_{\text{MD}}}[f(\boldsymbol{x}) \neq y]$. We actually use BSD as an auxiliary task and only care about the performance of MD.

\subsection{Model Details}
We present AdMul to tackle MD and BSD simultaneously. As Fig. \ref{model} shows, AdMul has five key parts: shared feature extractor $Q_f$ (BERT in our case, the green part), task-specific classifier $Q_y$ (the purple part), gradient reversal layer $Q_{\lambda}$ ( the grey part), global task discriminator $Q_d^g$ (the red part) and local task discriminators $Q_d^{l_c}$ (the yellow part).

\subsubsection{Feature Extractor}
AdMul adopts BERT as the feature extractor $Q_f$, which is shared by both MD and BSD. We take the BERT hidden state of [CLS] as a semantic summary of the input segment pair \citep{devlin-etal-2019-bert}. [CLS] can automatically learn the positions of two target words in the two segments, and then perceive the semantic difference via self-attention mechanism \citep{attention}. The hidden state then goes through a non-linear activation function and produces semantic discrepancy feature $\boldsymbol{v}$:
\begin{equation}
    \boldsymbol{v} = {\rm tanh}\left ( Q_f\left ( x_{[CLS]}\right ) \right ).
    \label{eq:f}
\end{equation}

On the other hand, we use the whole input sequence $\boldsymbol{x}$ to generate sentence embedding $\boldsymbol{h}$ via average 
 pooling:

 \begin{equation}
     \boldsymbol{h} =  Q_f \left ( \boldsymbol{x}\right ) .
     \label{eq:h}
 \end{equation}
 
\subsubsection{Task-specific Classifier}
Task-specific classifier $Q_y$ takes semantic discrepancy feature $\boldsymbol{v}$ as input. For the sake of brevity, we only draw a single classifier in the diagram. Actually, we are using use different classifiers for MD and BSD.
\begin{align}
    \boldsymbol{\hat{y}} &= Q_y(\boldsymbol{v}) = {\rm softmax}(W_{Q_y}\boldsymbol{v} + b_{Q_y}),
    \label{eq:cls}
\end{align}
where $\boldsymbol{\hat{y}} \in \mathbb{R}^2$ is the predicted label distribution of $\boldsymbol{x}$. $W_{Q_y}$ and $b_{Q_y}$ are weights and bias of $Q_y$. Finally, we can compute classification losses:

\begin{align}
    \mathcal{L}_y^{\rm MD} &= \frac{1}{{|\mathcal{D}_{\rm MD}|}}{\sum}_{i=1}^{|\mathcal{D}_{\rm MD}|} L_{CE}\left ( \hat{y}_i, y_i \right ),  \label{eq:md_loss}\\
    \mathcal{L}_y^{\rm BSD} &= \frac{1}{|\mathcal{D}_{\rm BSD}|}{\sum}_{i=1}^{|\mathcal{D}_{\rm BSD}|} L_{CE}\left ( \hat{y}_i, y_i \right ) \label{eq:bsd_loss},
\end{align}
where $L_{CE}$ is a cross-entropy loss function. $\hat{y}_i$ and $y_i$ are the predicted probability and the ground truth label of the i-th training sample respectively.

\subsubsection{Gradient Reversal Layer}
Gradient Reversal Layer (GRL) $Q_{\lambda}$ is the key point of adversarial learning \citep{grl}. During the forward propagation, GRL works as an identity function. While during the back propagation, it will multiply the gradient by a negative scalar $-\lambda$ to reverse the gradient. The operations can be formulated as the following pseudo function:
\begin{align}
    Q_{\lambda}(\boldsymbol{h}) &= \boldsymbol{h}, \\
    \frac{\partial Q_{\lambda}(\boldsymbol{h})}{\partial \boldsymbol{h}} &= -\lambda I, \label{eq:grl}
\end{align}
where $I$ is an identity matrix and $\lambda$ can be computed automatically (see Section \ref{implement}).

\subsubsection{Global Discriminator}
Sentence embedding $\boldsymbol{h} = Q_f(\boldsymbol{x})$ first goes through GRL, then global discriminator $Q_d^g$ tries to predict which task $\boldsymbol{h}$ belongs to. The training objective of $Q_d^g$ is:
\begin{align}
    \mathcal{L}_d^g =
     \frac{1}{|\mathcal{D}|}\sum_{
    \boldsymbol{x}_i \in {
    \mathcal{D}
    }
    }
    L_{CE} \left (Q_d^g(Q_f(\boldsymbol{x}_i)), d_i \right ),
\end{align}
where $\mathcal{D}=\mathcal{D}_{\rm MD} \cup \mathcal{D}_{\rm BSD}$. $d_i$ is the task label for input $\boldsymbol{x}_i$ ($d=0$ for MD and $d=1$ for BSD).

The feature extractor $Q_f$ tries to generate similar features to fool global task discriminator $Q_d^g$, so that $Q_d^g$ cannot accurately discern the source task of the input feature. The features that cannot be used to distinguish the source are task-invariant \citep{liu-etal-2017-adversarial, chen-cardie-2018-multinomial}. As the model converges, $Q_f$ will learn universal representations to align the distributions for MD and BSD.

\subsubsection{Local Discriminator}
\begin{figure}[H]
\centering
\scalebox{0.7}{
    \includegraphics[width=\columnwidth]{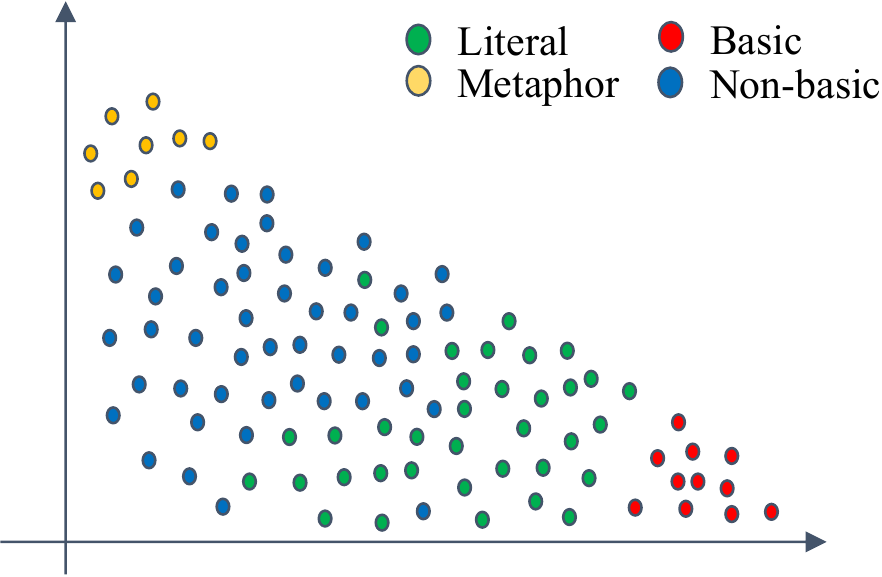}
}
\caption{Multi-mode structures of MD and BSD data distributions.}
\label{data_pattern}
\end{figure}

We noticed some corresponding patterns between MD and BSD via simple linguistic analysis. As Fig. \ref{data_pattern} illustrates. 

The samples in MD can be classified as literal or metaphorical, while the samples in BSD can be categorized as basic or non-basic. A basic sense (red samples) must be literal (green samples), so they are clustered closer in the feature space. A metaphor (yellow samples) must be non-basic (blue samples), hence they are closer. Moreover, the metaphorical and the basic are significantly dissimilar, so they lie at different corners in the feature space, far from each other. If we bring the literal and the basic closer, then the dividing line between the metaphorical and the literal will be clearer. If the metaphorical and the non-basic get closer, then BSD will be promoted as well. Better performance of BSD will strengthen knowledge transfer from BSD to MD.

Such multi-mode patterns inspire us to apply fine-grained alignment \citep{mingsheng, wang-2019}. We forcibly push the class 0 samples (literal in MD and basic in BSD) closer, and cluster the class 1 samples (metaphor in MD and non-basic in BSD) closer. Therefore, we use two local discriminators. Each aligns samples from class $c \in \{0, 1\}$:
\begin{align}
    \mathcal{L}_d^{l}=\frac{1}{|\mathcal{D}|} \sum_{c=0}^C \sum_{\boldsymbol{x_i} \in {
    \mathcal{D}}} w_d L_{CE} ( Q_d^{l_c} (\hat{y}_i^cQ_f(\boldsymbol{x}_i)), d_i ),
\label{eq:local}
\end{align}

where $d_i$ is the task label and $C$ is the number of classes. $d=0$ for MD and $d=1$ for BSD. $w_d$ is a task weight. To maintain the dominance of MD in local alignment, we set $w_0=1$ and $w_1=0.3$ in all experiments. $\hat{y}_i^c$ comes from Eq. \ref{eq:cls}. The classifier $Q_y$ will deliver a normalized label distribution for each sample $\boldsymbol{x}_i$, no matter which task it belongs to. We can view it as an attention mechanism. $Q_y$ thinks $\boldsymbol{x}_i$ has a probability of $\hat{y}_i^c$ to be class $c$. Then we use the label distribution as attention weights to apply to the sample. In practice, it performs better than hard attention, because more information can be considered.

The training of local discriminators is also adversarial. The feature extractor $Q_f$ generates task-invariant features to fool local discriminators $Q_d^{l_c}$, so that $Q_d^{l_c}$ cannot discern which task the features in class $c$ come from.

\subsubsection{Training Objective}
The training of AdMul involves multiple objectives. It can be formulated as the loss function below:
\begin{align}
    \mathcal{L} (& \theta_f, \theta_d, \theta_y) = \nonumber \\
     & \mathcal{L}_y^{\rm MD} +  \alpha \mathcal{L}_y^{\rm BSD} - \lambda ( \beta \mathcal{L}_d^g + \gamma \mathcal{L}_d^l ),
\label{eq:loss}
\end{align}
where $\alpha$, $\beta$ and $\gamma$ are hyper-parameters to balance the loss magnitudes. $\theta_f$, $\theta_d$ and $\theta_y$ are parameters of $Q_f$, $Q_d$ (all discriminators) and $Q_y$ respectively.

The optimization of $\mathcal{L}$ involves a mini-max game like Generative Adversarial Network \citep{gan}. The feature extractor $Q_f$ tries to make the deep features as similar as possible, so that both global and local task discriminators cannot differentiate which task they come from. After the training converges, the parameters $\hat{\theta}_f$, $\hat{\theta}_y$ and $\hat{\theta}_d$ will deliver a saddle point of Eq. \ref{eq:loss}:
\begin{align}
    (\hat{\theta}_f, \hat{\theta}_y) = \arg \min_{\theta_f, \theta_y}  \mathcal{L}(\theta_f, \theta_y, \theta_d),  \\
    (\hat{\theta}_d) = \arg \max_{\theta_d}  \mathcal{L}(\theta_f, \theta_y, \theta_d).
\label{eq:total_loss}
\end{align}

At the saddle point, $\theta_y$ will minimize classification loss $\mathcal{L}_y$ (combined by $\mathcal{L}_y^{\rm MD}$ and $\mathcal{L}_y^{\rm BSD}$). $\theta_d$ will minimize task discrimination loss $\mathcal{L}_d$ (combined by $\mathcal{L}_d^g$ and $\mathcal{L}_d^l$). $\theta_f$ will maximize the loss of task discriminators (features are task-invariant, so the task discrimination loss increases). AdMul can be easily trained via standard gradient descent algorithms. We take stochastic gradient descent (SGD) as an example:
\begin{align}
    \theta_f &\longleftarrow \theta_f - \eta \left ( \frac{\partial \mathcal{L}_y^i}{\partial \theta_f} - \lambda \frac{\partial \mathcal{L}_d^i}{\partial \theta_f} \right ), \\
    \theta_y &\longleftarrow \theta_y - \eta \left ( \frac{\partial \mathcal{L}_y^i}{\partial \theta_y} \right ), \\
    \theta_d &\longleftarrow \theta_d - \eta \left ( \frac{\partial \mathcal{L}_d^i}{\partial \theta_d} \right ),
\end{align}
where $i$ denotes the i-th training sample and $\eta$ is learning rate. The update for $\theta_y$ and $\theta_d$ is the same as SGD. As for $\theta_f$, if there is no minus sign for $\frac{\partial \mathcal{L}_d^i}{\partial \theta_f}$, then SGD will minimize the task discrimination loss $\mathcal{L}_d$, which means the features generated by $Q_f$ are dissimilar across tasks \citep{grl}. 

\section{Experiments}
\subsection{Datasets}
Four metaphor detection datasets are used in our experiments. The information is shown in Table \ref{tab:md_data}.

\begin{table}[ht]
\centering
\scalebox{0.75}{
\begin{tabular}{ccccc}
\toprule
\textbf{Dataset} & \textbf{\#Sent.} & \textbf{\#Tar.} & \textbf{\%Met.} & \textbf{Avg. Len} \\ \midrule
VUA $\operatorname{All}_{tr}$  & 6,323 & 116,622 & 11.19 & 18.4 \\
VUA $\operatorname{All}_{val}$ & 1,550 & 38,628 & 11.62 & 24.9 \\
VUA $\operatorname{All}_{te}$ & 2,694 & 50,175 & 12.44 & 18.6 \\ \midrule
VUA $\operatorname{Verb}_{tr}$ & 7,479 & 15,516 & 27.90 & 20.2 \\
VUA $\operatorname{Verb}_{val}$ & 1,541 & 1,724 & 26.91 & 25.0 \\
VUA $\operatorname{Verb}_{te}$ & 2,694 & 5,873 & 29.98 & 18.6 \\ \midrule
MOH-X & 647 & 647 & 48.69 & 8.0\\ \midrule
TroFi & 3,737 & 3,737 & 43.54 & 28.3\\ 
\bottomrule
\end{tabular}
}
\caption{MD Datasets information. \textbf{\#Sent.}: Number of sentences. \textbf{\#Tar.}: Number of target words. \textbf{\%Met.}: Proportion of metaphors. \textbf{Avg. Len}: Average sentence length.}
\label{tab:md_data}
\end{table}

\noindent \textbf{VUA All} \citep{steen2010method} is the largest metaphor detection dataset to date. VUA All labels each word in a sentence. The sentences are from four genres, namely academic, conversation, fiction, and news. \textbf{VUA Verb} \citep{steen2010method} is drawn from VUA All dataset. The target words are all verbs. \textbf{MOH-X} \citep{mohammad-etal-2016-metaphor} is sampled from WordNet, with only verb targets included. WordNet creates a sense inventory for each verb, of which some may have metaphorical senses. \textbf{TroFi} \citep{birke-sarkar-2006-clustering, birke-sarkar-2007-active} is a dataset collected from 1987-1989 Wall Street Journal Corpus via an unsupervised method. TroFi only has verb targets as well.

We use a word sense disambiguation (WSD) toolkit \citep{wsd_toolkit} to create the basic sense discrimination (BSD) dataset. The toolkit provides SemCor \citep{semcor}, the largest manually annotated dataset for WSD. We filter out the targets that have less than 3 senses to balance the magnitudes of WSD and MD datasets. The information of BSD dataset is shown in Table \ref{tab:bsd_data}.

\begin{table}[ht]
\centering
\scalebox{0.8}{
\begin{tabular}{ccccc}
\toprule
\textbf{Dataset} & \textbf{\#Sent.} & \textbf{\#Tar.} & \textbf{\%Basic} & \textbf{Avg. Len} \\ \midrule
$\text{SemCor}_{\text{BSD}}$ & 34,479 & 130,808 & 60.83 & 22.34 \\
\bottomrule
\end{tabular}
}
\caption{BSD Dataset information.  \textbf{\%Basic}: Proportion of basic senses. }
\label{tab:bsd_data}
\end{table}

\subsection{Compared Methods}
\noindent \textbf{RNN\_ELMo} and \textbf{RNN\_BERT} \citep{gao-etal-2018-neural} are two end-to-end models use both GloVe and ELMo embeddings. 

\noindent \textbf{RNN\_HG} and \textbf{RNN\_MHCA} \citep{mao-etal-2019-end} are based on RNN. Both models regard the static GloVe embedding as literal, and dynamic ELMo embedding can present metaphorical senses. RNN\_HG and RNN\_MHCA also utilize linguistic rules. 

\noindent \textbf{MUL\_GCN} \citep{Le_Thai_Nguyen_2020} uses multi-task learning to transfer knowledge from WSD to MD. However, it does not use shared layers. The knowledge transfer is accomplished via a loss term. MUL\_GCN also leverages dependency relations. 

\noindent \textbf{DeepMet} \citep{su-etal-2020-deepmet} is the winning method in the 2020 VUA and TOEFL Metaphor Detection Shared Task \citep{leong-etal-2020-report}. DeepMet is built upon BERT, with various external resources like fine-grained part of speech tags utilized. 

\noindent \textbf{MelBERT} \citep{choi-etal-2021-melbert} is designed upon RoBERTa. It uses a late-interaction mechanism to encode the literal meaning and the contextual meaning of a target respectively. MelBERT also leverages part of speech information. 

\noindent \textbf{MrBERT} \citep{song-etal-2021-verb} uses relation classification paradigm to detect metaphors. It embeds dependency relations into input to fine-tune pre-trained BERT, with various relation models applied.  

\noindent \textbf{MisNet} \citep{zhang-liu-2022-metaphor} is a linguistics-driven model. Two linguistic rules, namely Metaphor Identification Procedure and Selectional Preference Violation \citep{wilks1975preferential, wilks1978making} guide the model design. MisNet regards MD as semantic matching, with dictionary resources leveraged.

\begin{table*}[ht]
\renewcommand\thetable{4} 
\centering
\resizebox{\textwidth}{!}{
\begin{tabular}{l|cccc|cccc|cccc|cccc}
\toprule
\multicolumn{1}{c|}{\multirow{2}{*}{\textbf{Model}}} & \multicolumn{4}{c|}{\textbf{VUA All}} & \multicolumn{4}{c|}{\textbf{VUA Verb}} & \multicolumn{4}{c|}{\textbf{MOH-X (10 fold)}} &  \multicolumn{4}{c}{\textbf{TroFi (10 fold)}} \\
\multicolumn{1}{c|}{} & Pre. & Rec. & F1 & Acc. & Pre. & Rec. & F1 & Acc. & Pre. & Rec. & F1 & Acc. & Pre. & Rec. & F1 & Acc. \\ \midrule
RNN\_ELMo & 71.6 & 73.6 & 72.6 & 93.1 & 68.2 & 71.3 & 69.7 & 81.4 & 79.1 & 73.5 & 75.6 & 77.2 & 70.1 & 71.6 & 71.1 & 74.6\\
RNN\_BERT & 71.5 & 71.9 & 71.7 & 92.9 & 66.7 & 71.5 & 69.0 & 80.7 & 75.1 & 81.8 & 78.2 & 78.1 & 70.3 & 67.1 & 68.7 & 73.4 \\
RNN\_HG & 71.8 & 76.3 & 74.0 & 93.6 & 69.3 & 72.3 & 70.8 & 82.1 & 79.7 & 79.8 & 79.8 & 79.7 & 67.4 & {\ul \textit{77.8}} & 72.2 & 74.9 \\
RNN\_MHCA & 73.0 & 75.7 & 74.3 & 93.8 & 66.3 & {\ul \textit{75.2}} & 70.5 & 81.8 & 77.5 & 83.1 & 80.0 & 79.8 & 68.6 & 76.8 & 72.4 & 75.2 \\ \midrule
MUL\_GCN & 74.8 & 75.5 & 75.1 & 93.8 & 72.5 & 70.9 & 71.7 & 83.2 & 79.7 & 80.5 & 79.6 & 79.9 & \textbf{73.1} & 73.6 & {\ul \textit{73.2}} & {\ul \textit{76.4}} \\ 
DeepMet & {\ul \textit{82.0}} & 71.3 & 76.3 & - & {\ul \textit{79.5}} & 70.8 & 74.9 & - & - & - & - & -  & - & - & - & -\\
MelBERT & 80.1 & 76.9 & 78.5 & - & 78.7 & 72.9 & 75.7 & - & - & - & - & - & - & - & - & - \\
MrBERT & \textbf{82.7} & 72.5 & 77.2 & {\ul \textit{94.7}} & \textbf{80.8} & 71.5 & {\ul \textit{75.9}} & {\ul \textit{86.4}} & 80.0 & {\ul \textit{85.1}} & 82.1 & 81.9 & 70.4 & 74.3 & 72.2 & 75.1 \\
MisNet & 80.4 & {\ul \textit{78.4}} & \textbf{79.4} & \textbf{94.9} & 78.3 & 73.6 & {\ul \textit{75.9}} & 86.0 & {\ul \textit{84.2}} & 84.0 & {\ul \textit{83.4}} & {\ul \textit{83.6}} & 67.5 & 77.6 & 71.9 & 73.6 \\ \midrule
AdMul & 78.4 & \textbf{79.5} & {\ul \textit{79.0}} & {\ul \textit{94.7}} & 78.5 & \textbf{78.1} & \textbf{78.3} & \textbf{87.0} & \textbf{87.4} & \textbf{88.8} & \textbf{87.9} & \textbf{88.0} & {\ul \textit{70.5}} &  \textbf{79.8} & \textbf{74.7} & \textbf{76.5} \\
\bottomrule
\end{tabular}
}
\caption[Caption for LOF]{MD Results on VUA All, VUA Verb, MOH-X, and TroFi. The first four baseline models are end-to-end. The best performance for each metric in \textbf{bold}, and the second best in \textit{\underline{italic underlined}}.}
\label{main_result}
\end{table*}

\begin{table*}[ht]
\renewcommand\thetable{5}
\centering
\resizebox{\textwidth}{!}{
\begin{tabular}{l|cccc|cccc|cccc|cccc}
\toprule
\multicolumn{1}{c|}{\multirow{2}{*}{\textbf{Model}}}  & \multicolumn{4}{c|}{\textbf{Verb}} & \multicolumn{4}{c|}{\textbf{Adjective}} & \multicolumn{4}{c|}{\textbf{Adverb}} & \multicolumn{4}{c}{\textbf{Noun}} \\
 & Pre. & Rec. & F1 & Acc. & Pre. & Rec. & F1 & Acc. & Pre. & Rec. & F1 & Acc. & Pre. & Rec. & F1 & Acc. \\ \midrule
RNN\_ELMo & 68.1 & 71.9 & 69.9 & - & 56.1 & 60.6 & 58.3 & - & 67.2 & 53.7 & 59.7 & 94.8 & 59.9 & 60.8 & 60.4 & - \\
RNN\_BERT & 67.1 & 72.1 & 69.5 & 87.9 & 58.1 & 51.6 & 54.7 & 88.3 & 64.8 & 61.1 & 62.9 & 94.8 & 63.3 & 56.8 & 59.9 & 88.6 \\
RNN\_HG & 66.4 & 75.5 & 70.7 & {\ul \textit{88.0}} & 59.2 & {\ul \textit{65.6}} & 62.2 & 89.1 & 61.0 & 66.8 & 63.8 & 94.5 & 60.3 & 66.8 & 63.4 & 88.4 \\
RNN\_MHCA & 66.0 & 76.0 & 70.7 & 87.9 & 61.4 & 61.7 & 61.6 & 89.5 & 66.1 & 60.7 & 63.2 & {\ul \textit{94.9}} & 69.1 & 58.2 & 63.2 & 89.8 \\ \midrule
DeepMet & \textbf{78.8} & 68.5 & 73.3 & - & \textbf{79.0} & 52.9 & 63.3 & - & {\ul \textit{79.4}} & 66.4 & 72.3 & - & {\ul \textit{76.5}} & 57.1 & 65.4 & - \\
MelBERT & 74.2 & 75.9 & {\ul \textit{75.1}} & - & 69.4 & 60.1 & 64.4 & - & \textbf{80.2} & 69.7 & \textbf{74.6} & - & 75.4 & 66.5 & {\ul \textit{70.7}} & - \\
MisNet & {\ul \textit{77.5}} & {\ul \textit{77.7}} & \textbf{77.6} & \textbf{91.4} & 68.8 & 65.2 & {\ul \textit{67.0}} & {\ul \textit{91.2}} & 76.4 & {\ul \textit{70.5}} & 73.3 & \textbf{96.3} & 74.4 & {\ul \textit{67.2}} & 70.6 & {\ul \textit{91.6}} \\ \midrule
AdMul & 77.2 & \textbf{78.1} & \textbf{77.6} & \textbf{91.4} & {\ul \textit{72.4}} & \textbf{66.9} & \textbf{69.5} & \textbf{92.0} & 76.3 & \textbf{71.3} & {\ul \textit{73.7}} & \textbf{96.3} & \textbf{77.0} & \textbf{70.3} & \textbf{73.5} & \textbf{92.4} \\
\bottomrule
\end{tabular}
}
\caption{MD Breakdown results on VUA All for open word classes, the most important parts of metaphor detection. The first four baseline models are end-to-end.}
\label{vua_break}
\end{table*}


\subsection{Implementation Details}
\label{implement}
We use $\text{DeBERTa}_{base}$ as the backbone (feature extractor $Q_f$ in Fig. \ref{model}) for all experiments \citep{deberta}, through the APIs provided by HuggingFace \citep{huggingface}. The embedding dimension is 768. We set the maximum input sequence length as 150. The optimizer is AdamW \citep{peters-etal-2019-tune}. We let $\alpha=0.2$, $\beta=0.1$, and $\gamma=0.1$ according to the model performance on VUA Verb, and apply them to the rest datasets. The total training epoch, batch size, and learning rate are specific for each dataset, as Table \ref{tab:hyper} shows. 

\begin{table}[H]
\renewcommand\thetable{3} 
\centering
\scalebox{0.8}{
\begin{tabular}{cccc}
\toprule
\textbf{Dataset} & \textbf{Epochs} & \textbf{Batch Size} & \textbf{LR}  \\ \midrule
VUA All & 8 & 64 & 3e-5 \\
VUA Verb & 5 & 64 & 3e-5 \\
MOH-X & 5 & 32 & 2e-5 \\
TroFI & 10 & 64 & 1e-5 \\
\bottomrule
\end{tabular}
}
\caption{Hyper-parameters. \textbf{LR} stands for learning rate.}
\label{tab:hyper}
\end{table}

Instead of using a fixed constant, the parameter $\lambda$ in GRL (Eq. \ref{eq:grl}) is set by $\lambda=\frac{m}{1+\exp(-10p)}-n$, where $m=1.4$ and $n=0.6$. $p=\frac{t}{T}$, where $t$ and $T$ are the current training step and the maximum training step respectively. $\lambda$ is increasing from 0.1 to 0.8 in our case. Such a method stabilizes the training \citep{grl}. At the beginning of the training, $\lambda$ should be small so that the generated feature is not too hard for task discrimination. With training going on, adversarial training can be strengthened for better knowledge transfer. We choose the best model on the validation set for testing. Since MOH-X and TroFi do not have the training, validation, and testing split, we leverage 10-fold cross-validation. In each iteration, we pack MD and BSD samples into a mini-batch input. They have the same amount (half of the batch size). All experiments are done on an RTX 3090 GPU and CUDA 11.6.

\section{Metaphor Detection Results}

\subsection{Overall Results}
To be consistent with previous studies \citep{mao-etal-2018-word, choi-etal-2021-melbert, zhang-liu-2022-metaphor}, we mainly focus on the F1 score. As Table \ref{main_result} shows, our proposed AdMul obtains great improvements compared with the baseline models. Best scores are reported on 3 out of 4 datasets, including VUA Verb, MOH-X, and TroFi. We attain a comparable result to the state-of-the-art model on VUA All as well. The average F1 score across 4 datasets is 79.98, which is 2.33 points higher than MisNet (77.65 on average). We notice that AdMul performs better on small datasets (VUA Verb, MOH-X, and TroFi) than the large dataset (VUA All). We attribute it to different dataset sizes. Deep learning models need numerous data to achieve good performance, so MTL can help. The knowledge distilled from BSD can greatly promote MD, especially when faced with severe data scarcity problems. MTL also works as a regularization method to avoid overfitting via learning task-invariant features \citep{liu-etal-2019-multi}. However, VUA All is a large dataset, so there may be a marginal utility for more data from a related task. VUA All requires predictions for each word class as well, while BSD only has open class (i.e., verb, noun, adjective, and adverb) words. Consequently, the rest word class targets cannot get enough transferred knowledge. 

The most significant enhancement is from MOH-X. BSD dataset and MOH-X are both built upon WordNet, so the data distributions can be very similar. In such a case, AdMul can easily align globally, and pay more attention to local alignment. The improvement from TroFi is barely satisfactory. TroFi is built via an unsupervised method, therefore it may contain many noises. Many baseline models perform mediocrely on TroFi as observed.

MUL\_GCN is the only chosen baseline method in our experiments. MUL\_GCN used an L2 loss term to force the encoder of MD and the encoder of WSD to generate similar deep features for both MD and WSD data. However, MUL\_GCN only leveraged the features at the output layer, without using parameter-sharing strategy. Thus MUL\_GCN did not allow latent interaction between different data distributions, and that is why our method performs better. 

\subsection{VUA All Breakdown Results}
Table \ref{vua_break} shows a breakdown analysis of VUA All dataset. The most important part of MD is the model performance on open class words. As we can see, AdMul achieves the best F1 scores on 3 out of 4 word classes, and acquires a result similar to MelBERT on adverbs. The biggest gains are reported on nouns, with 2.8 absolute F1 score improvements against the strongest baseline MelBERT. The enhancement in adjectives is also encouraging (2.5 absolute improvements against MisNet). Though AdMul performs slightly less well than MisNet on VUA All, AdMul obtains better results on open class words. As we mentioned before, WordNet only has annotated knowledge for open class words, which demonstrates that AdMul can get benefits from MTL.

\subsection{VUA All Genres}
\label{sec:genres}
The sentences of VUA All dataset originate from four genres, namely academic, conversation, fiction, and news. The performance of our proposed AdMuL on the four genres is shown in Table \ref{vua:genre}.

\begin{table}[ht]
\centering
\resizebox{0.7\columnwidth}{!}{
\begin{tabular}{c|cccc}
\toprule
\textbf{Genre} & \textbf{Pre.} & \textbf{Rec.} & \textbf{F1} & \textbf{Acc.} \\ \midrule
Academic & 83.9 & 83.5 & 83.7 & 94.4 \\
Conversation & 66.6 & 73.9 & 70.1 & 95.2 \\ 
Fiction & 74.5 & 81.7 & 77.9 & 95.7 \\
News & 81.1 & 76.4 & 78.7 & 93.7 \\
\bottomrule
\end{tabular}
}
\caption{Performance of four genres in VUA All. }
\label{vua:genre}
\end{table}

The performance of conversation is inferior to the others. Conversations have more closed word classes (e.g., conjunctions, interjections, prepositions, etc.). The performance on academic is the best, since it has more open class words, which are adequate in WordNet. VUA All dataset annotates metaphoricity for closed word classes as well. However, these cases may be confusing. 

e.g. \textit{She checks her appearance \textbf{in} a mirror.}

The preposition \textbf{\textit{in}} in the above sentence is tagged as metaphorical. However, it is quite tricky even for humans to notice the metaphorical sense. As Table \ref{vua:pos} shows, there are lots of words in closed classes, but our proposed AdMuL cannot get transferred knowledge from auxiliary task BSD.

\begin{table}[ht]
\centering
\resizebox{0.6\columnwidth}{!}{
\begin{tabular}{cc|ccc}
\toprule
 & \textbf{POS} & \textbf{Train} & \textbf{Val} & \textbf{Test} \\ \midrule
\multirow{4}{*}{\rotatebox{90}{\textbf{Open}}} & VERB & 20,917 & 7,152 & 9,872 \\
 & NOUN & 20,514 & 6,859 & 8,588 \\
 & ADJ & 9,673 & 3,213 & 3,965 \\
 & ADV & 6,973 & 2,229 & 3,393 \\ \midrule
\multirow{6}{*}{\rotatebox{90}{\textbf{Closed}}} & PART & 2,966 & 1,137 & 1,463 \\
 & PRON & 6,942 & 2,230 & 3,955 \\
 & ADP & 13,310 & 4,556 & 5,300 \\
 & DET & 10,807 & 3,541 & 4,118 \\
 & CCONJ & 3,645 & 1,369 & 1,581 \\
 & INTJ & 734 & 159 & 398 \\ \bottomrule
\end{tabular}
}
\caption{Number for different word classes in VUA All dataset.}
\label{vua:pos}
\end{table}

\subsection{Zero-shot Transfer}
We use AdMul trained on VUA All to conduct zero-shot transfer on two small datasets, i.e., MOH-X and TroFi. The results are shown in Table \ref{zero-shot}. Though the performance on VUA All is inferior to MisNet, AdMul has a stronger generalization ability, defeating the baseline models in all metrics across two datasets. It is worth mentioning that DeepMet and MelBERT are trained on an expanded version of VUA All \citep{choi-etal-2021-melbert}, so they have more data than us. Our zero-shot performance on MOH-X is even better than fine-tuned MisNet, the previous state-of-the-art method (see Table \ref{main_result}).

\begin{table}[ht]
\centering
\resizebox{\columnwidth}{!}{
\begin{tabular}{l|cccc|cccc}
\toprule
\multicolumn{1}{c|}{\multirow{2}{*}{\textbf{Model}}} & \multicolumn{4}{c|}{\textbf{MOH-X (Zero-shot)}} & \multicolumn{4}{c}{\textbf{TroFi (Zero-shot)}} \\
\multicolumn{1}{c|}{} & Pre. & Rec. & F1 & Acc. & Pre. & Rec. & F1 & Acc. \\ \midrule
DeepMet & {\ul \textit{79.9}} & 76.5 & 77.9 & - & 53.7 & 72.9 & 61.7 & - \\
MelBERT & 79.3 & 79.7 & 79.2 & - & 53.4 & 74.1 & 62.0 & - \\
MrBERT & 75.9 & 84.1 & 79.8 & 79.3 & {\ul \textit{53.8}} & 75.0 & 62.7 & 61.1\\ 
MisNet & 77.8 & {\ul \textit{84.4}} & {\ul \textit{81.0}} & {\ul \textit{80.7}} & {\ul \textit{53.8}} & {\ul \textit{76.2}} & {\ul \textit{63.1}} & {\ul \textit{61.2}}\\ \midrule
AdMul & \textbf{82.3} & \textbf{85.4} & \textbf{83.8} & \textbf{83.9} & \textbf{55.7} & \textbf{77.1} & \textbf{64.7} & \textbf{63.3}\\
\bottomrule
\end{tabular}
}
\caption{Zero-shot transfer results.}
\label{zero-shot}
\end{table}

\subsection{Ablation Study}
We carried out ablation experiments to prove the effectiveness of each module, as Table \ref{ablation} shows. We removed global discriminator $Q_d^g$, local discriminators $Q_d^{l_c}$, and adversarial training (no discriminators used) respectively. Each setting hurts the performance of the MTL framework. It demonstrates that we cannot naively apply MTL to combine MD and BSD. Instead, we should carefully deal with the alignment patterns globally and locally for better knowledge transfer. In addition, we tested $\text{DeBERTa}_{base}$, a model trained only on MD dataset. $\text{DeBERTa}_{base}$ takes the target word and its context as input, thus it can be viewed as a realization of MIP. The performance of $\text{DeBERTa}_{base}$ is mediocre, which indicates that the progress of AdMul is not only due to the large pre-trained language model, but closely related to our adversarial multi-task learning framework.

\begin{table}[htbp]
\centering
\resizebox{0.7\columnwidth}{!}{
\begin{tabular}{c|cccc}
\toprule
\textbf{Model} & \textbf{Pre.} & \textbf{Rec.} & \textbf{F1} & \textbf{Acc.} \\ \midrule
AdMul & \textit{\underline{78.5}} & \textit{\underline{78.1}} & \textbf{78.3} & \textbf{87.0} \\
\midrule
w/o global disc. & 75.0 & 77.3 & \textit{\underline{76.2}} & 85.5 \\ 
w/o local disc.  & 71.9 & \textbf{80.5} & 76.0 & 84.7 \\
w/o adv. & \textbf{79.3} & 73.0 & 76.0 & \textit{\underline{86.2}} \\
$\text{DeBERTa}_{base}$ & 78.2 & 71.3 & 74.6 & 85.4
\\ \bottomrule
\end{tabular}
}
\caption{Ablation on VUA Verb. w/o denotes \textit{without}. }
\label{ablation}
\end{table}

\subsection{Hyper-parameter Discussion}
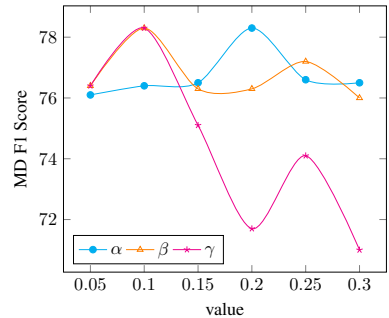
\begin{figure}[ht]
\centering
\scalebox{0.62}{
\begin{tikzpicture} 

\begin{axis}[
    xlabel=value, 
    ylabel=MD F1 Score, 
    legend columns=3,
    legend style={
        font=\mystrut,
        legend cell align=left,
    },
    legend pos=south west,
    ]

\addplot[smooth,mark=*,cyan] plot coordinates { 
    (0.05, 76.1)
    (0.1, 76.4)
    (0.15, 76.5)
    (0.2, 78.3)
    (0.25, 76.6)
    (0.3, 76.5)
};

\addlegendentry{$\alpha$}


\addplot[smooth,mark=triangle,orange] plot coordinates {
    (0.05, 76.4)
    (0.1, 78.3)
    (0.15, 76.3)
    (0.2, 76.3)
    (0.25, 77.2)
    (0.3, 76.0)
};
\addlegendentry{$\beta$}

\addplot[smooth,mark=star,magenta] plot coordinates {
    (0.05, 76.4)
    (0.1, 78.3)
    (0.15, 75.1)
    (0.2, 71.7)
    (0.25, 74.1)
    (0.3, 71.0)
};
\addlegendentry{$\gamma$}

\end{axis}
\end{tikzpicture}
}
\caption{Impacts of hyper-parameters.}
\label{fig:hyper}
\end{figure}

In Eq. \ref{eq:loss}, there are three hyper-parameters, i.e., $\alpha$, $\beta$, and $\gamma$ that balance the loss of BSD, global alignment loss, and local alignment loss respectively. Here we conduct experiments on VUA Verb dataset to see the impacts of different loss weight values. We tune each weight with the rest fixed. The results are shown in Fig. \ref{fig:hyper}. If $\alpha$ is too small, then the model cannot get enough transferred knowledge from BSD. On the contrary, if $\alpha$ is too large, then BSD will dominate the training, leading to poorer performance of MD. 

Two adversarial weights $\beta$ and $\gamma$ share the same pattern. If they are too small, then the data distributions cannot be aligned well globally or locally, resulting in inadequate knowledge transfer. On the contrary, if they are too big, distribution alignment will dominate the training. It is worth mentioning that the training is quite sensitive to $\gamma$, because our local alignment is based on a linguistic hypothesis. We should not pay much attention to local alignment, or it will disrupt the correct semantic space, leading to bad results.

\subsection{Hyper-parameter Search}
In this paper, the hyper-parameters are BSD loss weight $\alpha$, global alignment loss weight $\beta$, local alignment loss weight $\gamma$, learning rate $\eta$, batch size, and total training epoch. We tune each hyper-parameter with the rest fixed. $\alpha$, $\beta$, and $\gamma$ are searched from 0.05 to 0.5, with an interval of 0.05. $\eta$ is searched in $\left[ 1e-5, 2e-5, 3e-5, 4e-5, 5e-5 \right]$. The batch size is selected from $\left[ 16, 32, 64 \right]$. The total training epoch is selected from $\left[ 5, 8, 10 \right]$.
The best hyper-parameters are described in Section \ref{implement}. As mentioned before, we tune all hyper-parameters on VUA Verb dataset, and apply them to the rest datasets, except $\eta$, batch size, and the total training epoch.

\section{Conclusion}
In this paper, we proposed AdMul, an adversarial multi-task learning framework for end-to-end metaphor detection. AdMul uses a new task, basic sense discrimination to promote MD, achieving promising results on several datasets. The zero-shot results even surpass the previous fine-tuned state-of-the-art method. The ablation study demonstrates that the strong ability of AdMul comes not only from the pre-trained language model, but also from our adversarial multi-task learning framework.

\section*{Acknowledgement}
This work is supported by 2018 National Major Program of Philosophy and Social Science Fund (18ZDA238), and Tsinghua University Initiative Scientific Research Program (2019THZWJC38).

\section*{Limitations}
Though we simply assume that the most commonly used lexical sense is a more basic sense and such an assumption fits most cases, it may not be accurate all the time. Take the verb \textit{\textbf{dream}} as an example. The most commonly used sense of \textit{\textbf{dream}} according to WordNet is "have a daydream; indulge in a fantasy", which is metaphorical and non-basic. While it has another literal and basic sense, meaning "experience while sleeping". We are expecting a more fine-grained annotation system to clarify the evolution of different senses: which sense is basic and how other senses are derived. Such a system will benefit both metaphor detection and linguistic ontology studies.

Due to computing convenience, our model cannot handle long texts. An indirect metaphor needs to be determined across several sentences. Such a case is beyond our capabilities \citep{zhang-liu-2022-metaphor}. We will also leave it as a future work.

\section*{Ethics Statement}
Our proposed AdMul aims to detect metaphors in English, and the method can also be applied to other languages or multi-lingual cases. Though our manual observations did not show that there were biased metaphor detection cases for AdMul, there may still exist biases from the pre-trained language model.

We use $\text{DeBERTa}_{base}$ in all experiments, which is pre-trained on a variety of datasets, including Wikipedia, BookCorpus\footnote{https://github.com/butsugiri/homemade\_bookcorpus}, and CommonCrawl, etc\citep{deberta}. The total pre-training data size is about 78GB. Since AdMul needs to fine-tune $\text{DeBERTa}_{base}$, AdMul may inherit poisonous languages from the pre-trained language model, like hate speech, gender bias, stereotypes, etc.

\bibliography{anthology,custom}
\bibliographystyle{acl_natbib}

\appendix

\end{document}